\documentclass[10pt,twocolumn,letterpaper]{article}

\usepackage{cvpr}              % To produce the CAMERA-READY version
\usepackage{braket}
\usepackage{graphicx}
\usepackage{amsmath}
\usepackage{amssymb}
\usepackage{booktabs}
\usepackage{multirow}
\usepackage{multicol}
\usepackage{tabularx}
\usepackage{arydshln}
\usepackage{makecell}
\usepackage[T1]{fontenc}

\definecolor{cvprblue}{rgb}{0.21,0.49,0.74}
\usepackage[pagebackref,breaklinks,colorlinks,allcolors=cvprblue]{hyperref}

\def\paperID{*****} % *** Enter the Paper ID here
\def\confName{CVPR}
\def\confYear{2025}

\title{AV-EmoDialog: Chat with Audio-Visual Users \\ Leveraging Emotional Cues}

\author{Se Jin Park \quad\quad Yeonju Kim \quad\quad Hyeongseop Rha \quad\quad Bella Godiva \quad\quad Yong Man Ro$^\dagger$ \\
School of Electrical Engineering, KAIST, South Korea \\
{\tt\small \{jinny960812, yeonju7.kim, ryool\_1832, bgodiva, ymro\}@kaist.ac.kr}
}

\begin{document}
\maketitle
\begin{abstract}
In human communication, both verbal and non-verbal cues play a crucial role in conveying emotions, intentions, and meaning beyond words alone. These non-linguistic information, such as facial expressions, eye contact, voice tone, and pitch, are fundamental elements of effective interactions, enriching conversations by adding emotional and contextual depth. Recognizing the importance of non-linguistic content in communication, we present AV-EmoDialog, a dialogue system designed to exploit verbal and non-verbal information from users’ audio-visual inputs to generate more responsive and empathetic interactions. AV-EmoDialog systematically exploits the emotional cues in audio-visual dialogues; extracting speech content and emotional tones from speech, analyzing fine-grained facial expressions from visuals, and integrating these cues to generate emotionally aware responses in an end-to-end manner. Through extensive experiments, we validate that the proposed AV-EmoDialog outperforms existing multimodal LLMs in generating not only emotionally appropriate but also contextually appropriate responses.

\end{abstract}    
\section{Introduction}
\label{sec:intro}

Large language models (LLMs) \cite{touvron2023llama, brown2020language, anil2023palm, achiam2023gpt} excel in generating coherent and contextually appropriate text. Trained on extensive dialogue text, they facilitate human-machine interaction by generating relevant responses to user queries. However, the non-linguistic aspect of dialogue, crucial for effective and human-like communication has not been sufficiently explored. 

Human communication is inherently multi-modal, incorporating rich verbal and non-verbal cues. The non-linguistic cues, defined as any transfer of messages that do not involve the use of words, include facial expressions, eye movements, and paralinguistic features such as tone and pitch of voice \cite{urakami2023nonverbal, argyle1972non}. These cues convey emotions and intentions beyond words, significantly changing the meaning of a word. For example, the word "fine" when said with a cheerful tone and a smile, can genuinely mean everything is good, but when uttered with a flat tone and a stern face, it may imply dissatisfaction. This is the primary reason why many prefer face-to-face interactions over pure text-based ones to reduce potential misunderstandings. Therefore, the user's emotional state should be precisely and sufficiently considered to generate a contextually appropriate response.

Recent research \cite{alnuhait2023facechat, chu2024multimodalesc, fei2024empathyear} has attempted to address the emotional aspect of dialogue systems. Typically, these methods follow a cascade approach. They rely on text input, requiring a cascade of ASR module to obtain a transcription of spoken words \cite{alnuhait2023facechat,fei2024empathyear}. Furthermore, they employ a pretrained emotion classifier on either audio or visual \cite{abbasian2024empathy} to categorize into one of seven labels (\ie happy, sad, surprised, fearful, disgusted, angry, and neutral). However, relying on a single modality to extract emotional cues limits the richness of the emotional context, and categorizing emotion into seven labels simplifies the complex and subtle emotional states that naturally evolve during interactions. Such classification limits the system's ability to accurately learn the rich emotional nuances from the multimodal input. Moreoever, they are not optimized end-to-end across both audio and visual modalities, resulting in suboptimal emotional understanding and reduced efficiency in response generation.

In this paper, we introduce AV-EmoDialog, which directly processes audio-visual user input and generates an empathetic response based on the analyzed linguistic and non-linguistic information in end-to-end. We propose a systematic scheme to exploit fine-grained emotional cues in the audio-visual input by utilizing detailed descriptions relevant to the emotional context. The training proceeds in three stages. First, a speech encoder is trained to extract both speech content and nuanced emotion from the user's speech input through a high-performing chat-LLM. Second, a face encoder is trained to extract the fine-grained emotion from the user's facial video through the LLM. For fine-grained emotion extraction, we create detailed facial descriptions from the speaker face video. Last, it puts the speech and face encoders together with the LLM to directly aggregate the audio-visual information to generate emotionally resonant responses. Additionally, it generates the emotional state along with the response, enabling the model to consistently track and reflect the emotional tone throughout the conversation.

Our contributions are as follows: 
\begin{itemize}
\item{We present AV-EmoDialog, an end-to-end model that directly processes audio-visual user inputs to generate emotion-aware responses. This approach enables seamless interaction without relying on intermediate text.}
\item{We systematically optimize both the speech and face encoders to extract fine-grained verbal and non-verbal communicative signals. We find that leveraging detailed user metadata beyond emotion categories enhances emotion understanding in the dialogue.}
\item{Extensive experiments demonstrate that AV-EmoDialog excels at generating emotionally appropriate responses from audio-visual inputs. We validate the importance of exploiting both audio and visual modalities not only for emotional performance but also semantic performance of dialogue generation.}
\end{itemize}
\section{Related Works}
\label{sec:related_works}

\subsection{Spoken Dialogue Models}
Recent efforts have aimed to leverage the capabilities of large language models (LLMs) \cite{zhang2022opt,chowdhery2023palm,touvron2023llama,bai2023qwen} for enhancing spoken dialogue systems. They aim to autoregressively model the semantic and acoustic content of raw speech to process and generate speech. d-GSLM \cite{nguyen2023generative} models two-channel conversations to produce natural turn-taking conversations. SpeechGPT \cite{zhang2023speechgpt} initially converts speech to discrete speech tokens, followed by a three-stage training pipeline involving paired speech data, speech instruction data, and chain-of-modality instruction data. AudioGPT \cite{huang2023audiogpt} guides LLMs to generate commands for interacting with external tools before processing these commands within the models. Qwen-Audio-Chat \cite{chu2023qwen} is trained on extensive audio understanding tasks and instruction fine-tuned on dialogue framework. 

Some spoken models \cite{lin2024paralinguistics, xue2023chat, abbasian2024empathymc} have incorporated emotional cues from speech to enhance the model’s ability to tailor responses to diverse emotional contexts. By predicting emotion labels alongside each response, it helps the LLM better understand the user’s emotional state and enables it to generate responses that are more emotionally attuned and contextually appropriate. While these studies have discerned emotional cues within speech, they overlook non-verbal cues present in the visual modality. As human interaction is inherently multimodal, with visual cues such as facial expressions and eye contact playing a crucial role in conveying emotions and intentions, our work aims to integrate the audio and visual information of the user to better capture the communicative signals. 

\subsection{Multimodal Large Language Models} 
There has been a surge of Multimodal Large Language Models (MM-LLMs) \cite{li2023blip, liu2024visual, liu2024llava, maaz2023video, wu2023visual, shen2024hugginggpt, wu2023next, lee2024collavo, team2023gemini} that are capable of processing multimodal input. BLIP-2 \cite{li2023blip} introduces a Q-Former module to bridge the modality gap between the visual and text. Llava \cite{liu2024visual,liu2024llava} aligns the visual and text modality by utilizing a large-scale multimodal instruction-following dataset on image QA, image description, and complex reasoning. Video-LlaMA \cite{zhang2023video} and Video-ChatGPT \cite{maaz2023video} process the visual and audio content in a video to engage in a conversation describing the visual scenes and actions in a spatio-temporal context. HuggingGPT \cite{shen2024hugginggpt} conducts task planning to connect various AI models upon receiving a user request and NextGPT \cite{wu2023next} builds an end-to-end system that is capable of understanding and generating audio, visual, and text modality. However, they primarily concentrate on multimodal grounding tasks. They excel in integrating multimodal data for content reasoning but do not leverage this information to enhance interaction with the human user.

Recently, a few multimodal large language models \cite{alnuhait2023facechat, chu2024multimodalesc, fei2024empathyear} tailored for emotion-aware dialogue system attempted to incorporate emotional context. FaceChat \cite{alnuhait2023facechat} employs a pre-trained facial emotion classifier \cite{taigman2014deepface} to predict textual emotion labels from the user’s facial expression and incorporate this label into the LLM’s template prompt. It also needs an ASR module to transcribe speech into text, and utilizes the zero-shot capabilities of a pre-trained text LLM for response generation. SMES \cite{chu2024multimodalesc} utilizes a pre-trained Video-Llama \cite{zhang2023video} to extract emotion labels of each time step via prompting. These emotional labels are then combined with text inputs and fed into an LLM for response generation. EmpathyEar \cite{fei2024empathyear} utilizes an ImageBind \cite{girdhar2023imagebind} to extract multimodal embeddings, which are also combined with text embeddings to be fed into an LLM. Notably, these approaches follow a cascade approach, relying on pre-trained components such as emotion classifiers and ASR modules, which are not optimized end-to-end on audio-visual dialogue. Moreoever, they depend on emotion classifiers that assign emotions to one of only seven broad categories, restricting emotional understanding and leading to a simplified interpretation of temporally varying and complex user emotions. 

In contrast, our work presents an audio-visual spoken dialogue system that directly processes both linguistic and non-linguistic content from audio-visual input, without relying on intermediate text. Our system is trained end-to-end on audio-visual dialogue data, ensuring that the communicative signals from the input modalities are jointly learned. We demonstrate our proposed method's strong ability to generate responses that are both contextually and emotionally aligned with the user's current state, leading to more natural, empathetic interactions.

\begin{figure*}[t]
	\begin{minipage}[b]{\linewidth}
		\centering		
            \centerline{\includegraphics[width=17cm]{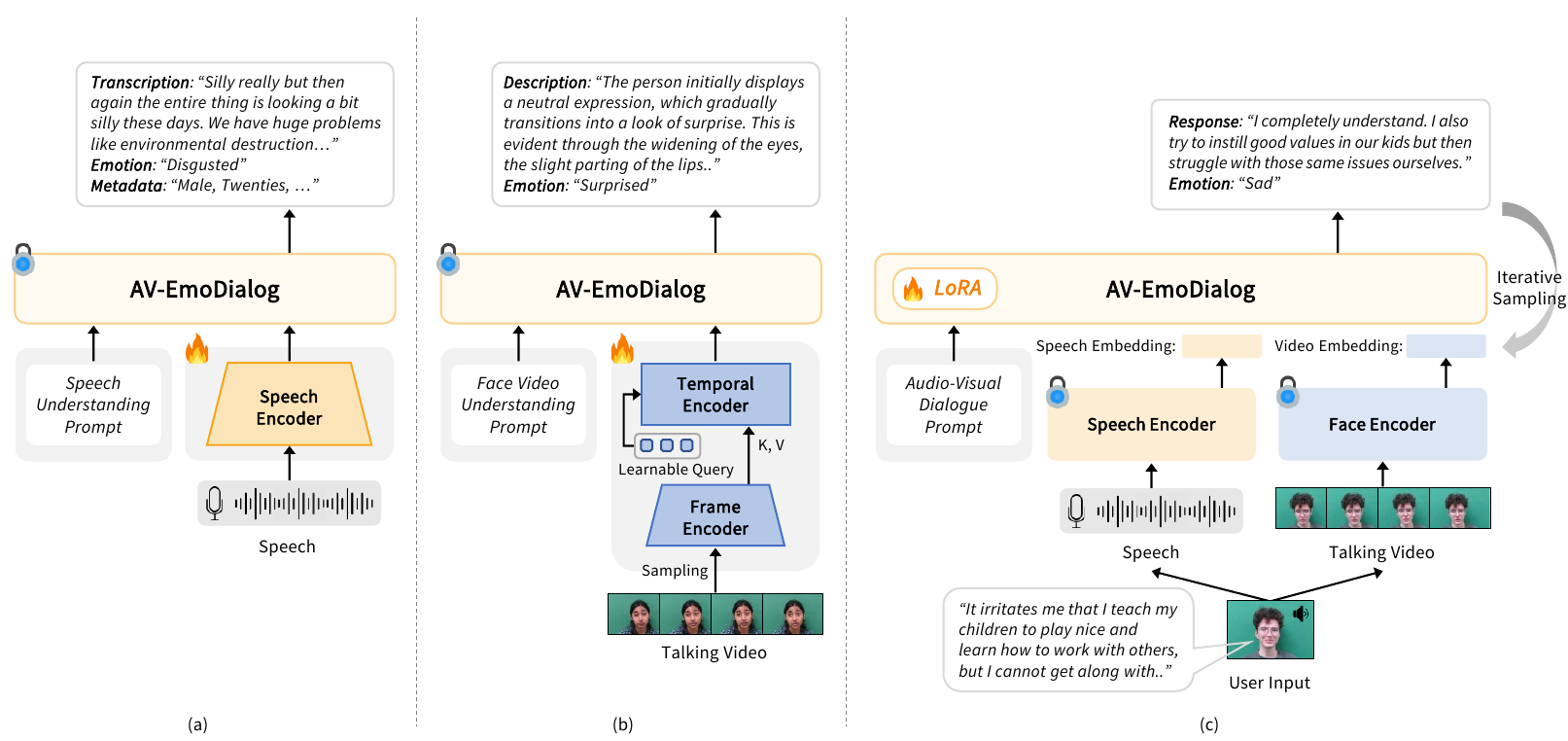}}
	\end{minipage}
        \vspace{-0.5cm}
	\caption{Overview of the proposed method where (a) the speech encoder is trained to extract verbal cues that can be understood by the LLM, (b) the face encoder is trained to extract nonverbal cues through the LLM, and (c) the LLM fine-tuned to integrate the verbal and nonverbal cues from the respective encoders to generate contextually and emotionally appropriate response.}
	\label{fig:1}
  \vspace{-0.2cm}
\end{figure*}
%##################################################
\section{Proposed Method}
\label{sec:method}

%#################################################
\begin{figure}[t]
	\begin{minipage}[b]{\linewidth}
		\centering		
            \centerline{\includegraphics[width=8cm]{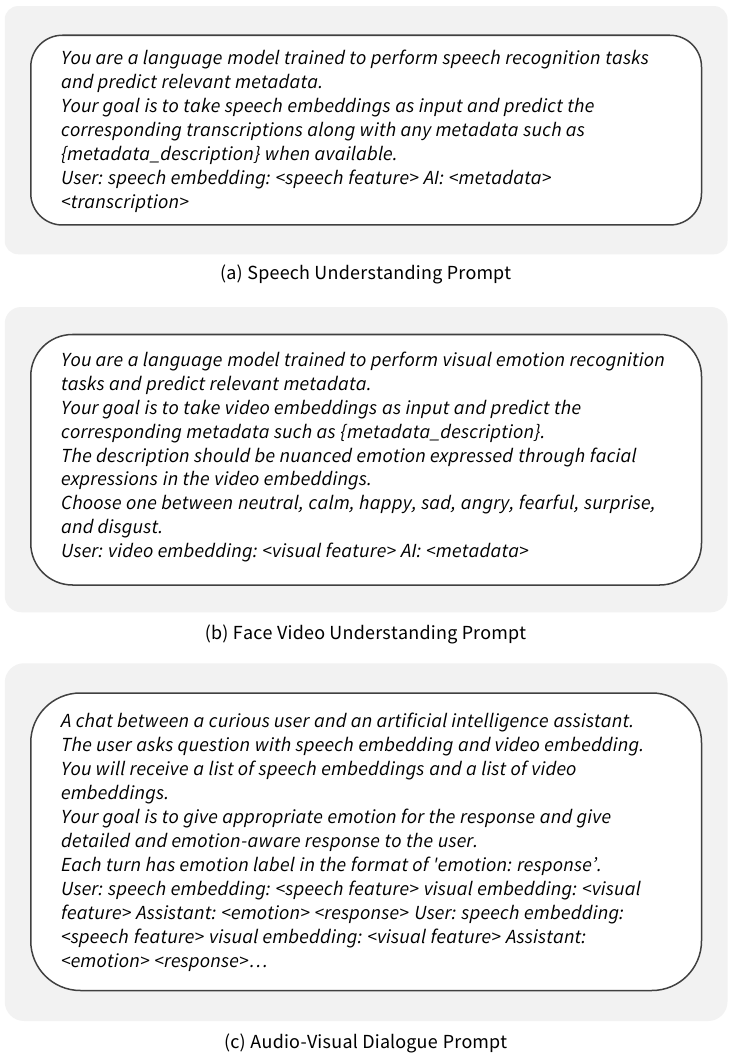}}
	\end{minipage}
        \vspace{-0.5cm}
	\caption{Hard prompts given at each stage of training. (a) speech understanding prompt enables the speech encoder to extract the verbal cues (both linguistic and non-linguistic information) from the audio input. (b) face video understanding prompt allows the face encoder to extract the nonverbal cues in the face video input. (c) audio-visual dialogue prompt guides the AV-EmoDialog to generate contextually and emotionally appropriate responses given the speech embedding and the visual embedding of the audio-visual user input. The metadata description is adjusted based on what extra information is available in the dataset such as emotion, emotion intensity, emotion description, age, and ethnicity.}
	\label{fig:4}
  \vspace{-0.2cm}
\end{figure}
%##################################################
Our approach to empowering large language models (LLMs) with the ability to interpret and respond to both verbal and non-verbal cues involves three stages: Sec \ref{sec:step1} extracting speech content verbal emotional cues from the audio input, Sec \ref{sec:step2} extracting non-verbal emotional cues from the user video input, and Sec \ref{sec:step3} integrating the verbal and non-verbal cues to build an audio-visual user engaged dialogue system. Below, we detail each stage of our methodology.

\subsection{Extracting Verbal Cues with Speech Encoder} 
\label{sec:step1}
From the raw audio input, we extract both the linguistic content (\ie what the speaker says) and verbal emotional cues. This stage equips an audio encoder with the ability to understand such communicative signals directly from speech such that the LLM can directly understand. We employ a strong audio encoder, Whisper model \cite{radford2023robust} which has shown powerful capabilities in speech-text processing tasks such as speech recognition (ASR), speech translation, voice activity detection. The audio encoder is connected to the high-performing chat-LLM, Llama-3-Instruct\footnote{\url{https://huggingface.co/meta-llama/Meta-Llama-3-8B-Instruct}\label{foot:llama3-Instruct}}, as illustrated in Figure \ref{fig:1}(a). We directly feed audio features $f_a$ from Whisper model to the LLM and align the feature space between the audio and text using an ASR training objective as in \cite{zhang2023speechgpt, chu2023qwen, wu2023next, tang2023salmonn} on extensive audio-text datasets; Gigaspeech \cite{chen2021gigaspeech}, CommonVoice \cite{ardila2019common}, and LibriSpeech \cite{panayotov2015librispeech}. By training on ASR objective, the model can directly understand the speech content and align the speech feature that can be interpreted by the LLM. 

Additionally, we incorporate speech emotion recognition (SER) objective to extract emotional tone from speech. By inputting speech features into the LLM instead of transcribed text, the LLM can infer deeper verbal aspects of speech such as the tone and pitch. We use speech data with emotion labels; RAVDESS \cite{livingstone2018ryerson}, CREMA-D \cite{cao2014crema}, MultiDialog \cite{park2024let}. 
The LLM remains frozen while we fine-tune the audio encoder. 
This dual training objective empowers the audio encoder to extract both linguistic content and emotional context that can be directly understood by the LLM. The hard prompt, namely the speech understanding prompt, given during this stage is shown in Figure \ref{fig:4}(a).

\subsection{Extracting Non-Verbal Cues with Face Encoder}
\label{sec:step2}
The face video contains rich non-verbal cues essential for nuanced human communication. To capture these subtle non-verbal cues, we deploy a face encoder with the ability to recognize the fine-grained facial emotion that can be directly interpreted by the LLM. As shown in Figure \ref{fig:1}(b), the face encoder consists of two components; the frame encoder and the temporal encoder. The frame encoder, CLIP-ViT \cite{radford2021learning}, known for strong visual-text alignment, extracts facial features at the frame level. Then, the temporal encoder aggregates the frame-level features to learn facial dynamics as emotion varies over time. Similar to \cite{jaegle2021perceiver,carion2020end}, we employ learnable queries that cross-attend to the frame-level features to output a fixed-length video feature $f_v$. This allows for the efficient processing of videos of varying lengths, which not only reduces computational costs but also facilitates handling longer dialogue sequences. 
The output visual features are directly fed into the LLM, and trained to recognize emotion labels on RAVDESS \cite{livingstone2018ryerson}, CREMA-D \cite{cao2014crema}, and MultiDialog \cite{park2024let} datasets. 

To optimize the training of the face encoder, we incorporate detailed descriptions of the facial expressions. Specifically, emotion-labeled face videos in the training datasets are further annotated with comprehensive descriptions of facial dynamics using GPT-4, as illustrated in Figure 2. These descriptions include granular details such as movements of the cheeks, lips, and eyes, as well as the progression of emotional expressions over time. By incorporating comprehensive facial descriptions along with the emotion labels as the training objectives, the face encoder learns to recognize the nuanced and dynamic aspects of human expressions that go beyond simplified emotion categories. Emotions during communication are inherently time-varying, often not consistently maintaining the same emotion. The time-considered detailed descriptions help the model recognize the evolving emotional states that go beyond the static and singular emotion categorization. 
It also guides to focus on specific facial attributes for accurate emotion detection, akin to how humans process emotion from the face. The face encoder is fine-tuned while the LLM remains frozen. The hard prompt, namely the face video understanding prompt, given to LLM during this stage is shown in Figure \ref{fig:4} (b). 

\subsection{Audio-Visual Dialogue Modeling} 
\label{sec:step3}
We build an emotion-aware audio-visual dialogue framework, namely AV-EmoDialog, by integrating the speech encoder, face encoder, and a LLM as shown in Figure \ref{fig:1} (c). Since the speech encoder and the face encoder are equipped with strong capabilities to process the linguistic and non-linguistic cues from audio-visual user input from the previous stages, the LLM is simply adapted to incorporate the multimodal communicative signals to generate contextually appropriate responses. In our setting, each dialogue consists of $R$ rounds of turns between the AI and the user, with the AI producing textual responses to audio-visual provided by the user. We use audio-visual dialogue language modeling with the following objective: 
\begin{align}
\mathcal{L}_{\text{dialog}} = \sum_{r=1}^{R} -\log p(e_{r}^{ai}, T_{r}^{ai} \mid f_{a, r}^{user}, f_{v, r}^{user},  \mathcal{H}), 
\end{align}
where $e_{r}^{ai}$. $T_{r}^{ai}$ are the emotional state and textual response of the AI at round $r$, which the LLM has to predict given the audio feature $f_{a, r}^{user}$ and visual feature $f_{v, r}^{user}$ of the user input at round $r$, and the dialogue history $\mathcal{H}$. 
We keep the audio and visual encoders frozen and LoRA \cite{hu2021lora} fine-tune the LLM on the target dialogue data, preserving the linguistic prowess while tailoring on the target audio-visual dialogue task. Through this focused adaptation, AV-EmoDialog not only responds accurately in text but also aligns its responses with the emotional and contextual nuances captured from the user's audio-visual cues. The hard prompt, audio-visual dialogue prompt is shown in Figure \ref{fig:4} (c). 

% In addition, we provide emotion-conditioned TTS, which can be used in conjunction with our AV-EmoDialog to build an audio-visual-to-audio dialogue system. Based on \cite{guo2023prompttts}, it extracts BERT embeddings from the emotion outputted by the dialogue system, which are concatenated with the content embeddings of the response to generate speech that conveys the correct textual information as well as the intended emotional tone. 

%#################################################
\begin{figure}[t]
	\begin{minipage}[b]{\linewidth}
		\centering		
            \centerline{\includegraphics[width=8cm]{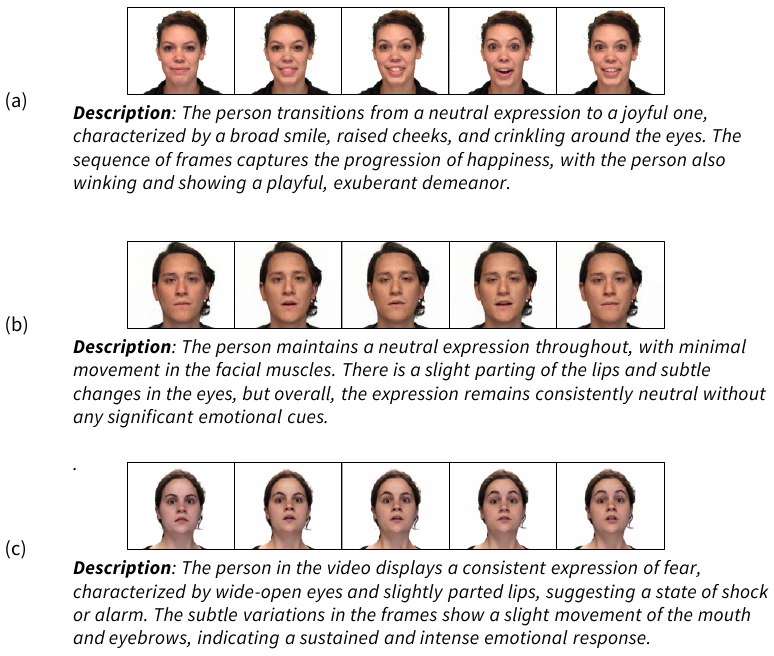}}
	\end{minipage}
        \vspace{-0.5cm}
	\caption{Examples of emotion-relevant descriptions  from face videos generated by GPT-4. The annotated descriptions provide detailed facial expressions and their dynamic changes over time, guiding the model to learn subtle and non-static emotion during the conversation}
	\label{fig:2}
  \vspace{-0.2cm}
\end{figure}
%##################################################

\section{Experiments}
\label{sec:experiment}

\subsection{Evaluation Metrics}
We evaluate the dialogue generation in terms of semantic quality and emotional quality. For the semantic quality, we employ standard metrics used for text-based dialogue generation: BLEU-1, BLEU-4 \cite{post2018call}, ROUGE \cite{lin2004rouge}, METEOR \cite{banerjee2005meteor}, and PPL \cite{bengio2000neural}. PPL is calculated using Dialog-GPT \cite{zhang2019dialogpt} and it is calculated across the test set. For the emotional quality, we measure EmoBERT score \cite{zhang2019bertscore} which calculates semantic similarity between the generated response and reference response using a BERT model finetuned on GoEmotion dataset \cite{demszky2020goemotions}. We additionally measure the diversity score with DISTINCT-1 \cite{li2015diversity} which quantifies how many unique unigrams (individual words) appear in the generated text relative to the total number of unigrams. For a more comprehensive evaluation, we additionally conduct GPT-4 evaluation and human evaluation. GPT-4 evaluation rates response given the dialogue history from 0 to 5 in terms of fluency and emotional intelligence. Likewise, human evaluation directly ranks the models in terms of contextual appropriateness and emotional intelligence. The detailed instructions for the evaluations are in the Appendix. 

\subsection{Implementation Details}
For the speech encoder, we adopt a pre-trained Whisper model  \cite{radford2023robust} of Qwen-Audio \cite{chu2023qwen} and fine-tune with ASR and speech emotion recognition objectives on a total of 6,485 hours of paired audio-text data, including Gigaspeech \cite{chen2021gigaspeech}, CommonVoice \cite{ardila2019common}, LibriSpeech \cite{panayotov2015librispeech}, RAVDESS \cite{livingstone2018ryerson}, CREMA-D \cite{cao2014crema}, and MultiDialog \cite{park2024let}. Note that we utilize the extra metadata available such as emotion, gender, age, and ethnicity to enrich the non-linguistic aspect of the speech feature. 

For the face encoder, we utilize a pre-trained CLIP-ViT as the frame encoder, followed by a transformer encoder with 6 layers and 8 attention heads as the temporal encoder. The videos are preprocessed into face crops of size 96$\times$96 using a face detector \cite{deng2020retinaface} and a facial landmark detector \cite{bulat2017far}. Each video is sampled every 10th frame to extract the frame-level features, which are further encoded into a fixed-length visual representation using the learnable queries of size 128. For obtaining the facial emotion description, we created a collage of images from the video frames as the GPT-4 is more adept at analyzing an image than a video. 

For the LLM, we use LlaMA3-Instruct\footref{foot:llama3-Instruct} which has a strong ability to understand and generate responses based on user instructions. In the first two stages, it is frozen, and in the last stage, we LoRA fine-tune \cite{hu2021lora} on the audio-visual dialogue data, MultiDialog \cite{park2024let}. Multidialog is a recently introduced large-scale multimodal dialogue corpus that includes audio, visual, and text modalities. All three stages are trained on 4 A6000 GPUs with AdamW optimizer and a cosine learning rate decay with a warm-up period. The lora rank is set to 16 and we enable bias and norm tuning. The max token length is 4096 with a batch size of 32. 
Please refer to the Appendix for the detailed statistics of the data.

\subsection{Baseline Methods}
We compare with the state-of-the-art multimodal language models that use various combinations of the modality. Llava-next (video) \cite{liu2024llava} takes a video and text as input to generate a response in text. SpeechGPT \cite{zhang2023speechgpt} takes speech input to generate a response in both text and speech. Since these baseline models are not tailored for dialogue generation tasks but instead on video understanding, captioning, speech recognition, and generation, we finetuned them on the testing dataset, MultiDialog \cite{park2024let}. We also construct a cascaded baseline by sequentially stacking state-of-the-art pre-trained models; audio speech recognition \cite{chu2023qwen}, speech emotion recognition \cite{chu2023qwen}, and Llama3 \footref{foot:llama3-Instruct}, where the audio model first transcribes the speech and recognizes the emotion which then goes into the LLM to generate emotion-aware response. We also finetune the cascaded baseline on the testing data for a fair comparison. 

\begin{table*}
	\renewcommand{\arraystretch}{1.2}
	\renewcommand{\tabcolsep}{2mm}
\centering
\resizebox{0.999\linewidth}{!}{
\begin{tabular}{cccccccccc}
\Xhline{3\arrayrulewidth}
\multirow{2}{*}{\textbf{Method}} & \multirow{2}{*}{\makecell{\textbf{Input-Output}\\ \textbf{Modality}}} &
\multicolumn{5}{c}{\textbf{Semantic}} & 
\multicolumn{1}{c}{\textbf{Emotion}} & 
\multicolumn{1}{c}{\textbf{Diversity}} \\ \cmidrule(lr){3-7} \cmidrule(lr){8-8} \cmidrule(lr){9-9} &

& \textbf{BLEU-1} & \textbf{BLEU-4} & \textbf{ROUGE} & \textbf{METEOR} & \textbf{PPL} & \textbf{EmoBERT}& \textbf{Dist-1} \\ \hline

% {LLaVA-NeXT-img \cite{liu2024llava}} & img/text-txt & 0.042 & 0.0006 & 0.019 & 0.050 & 1745.993 & 0.17 \\

{LLaVA-NeXT-vid \cite{liu2024llava}} & vid/txt-txt & 0.137 & 0.0135 & 0.150 & 0.112 & 317.410 & 0.21 & 0.873\\

{SpeechGPT \cite{zhang2023speechgpt}} & aud-txt/aud & 0.184 & 0.0145 & 0.192 & 0.163 & 285.654 & 0.14  & 0.876 \\

{Qwen-Audio \cite{chu2023qwen}+ llama3-Instruct} & aud-txt & 0.196 &0.0304 & \bf{0.226} & 0.154 & \bf{241.873} & 0.22 & 0.816 \\

\hdashline 
\multicolumn{4}{l}{\quad $\bullet$ \textbf{\textit{Proposed Method}}} \\ 
\textbf{AV-EmoDialog*} & aud/vid-txt & \bf{0.212} & 0.0272 &  0.215 & \bf{0.169} & 380.891 & 0.20 & 0.862 \\

% \textbf{AV-EmoDialog} & aud/vid-txt &  \bf{0.212} & \bf{0.0293} & \bf{0.216} & \bf{0.179} & 345.005 & \bf{0.21}   \\

\textbf{AV-EmoDialog} & aud/vid-txt &  0.193 & \bf{0.0307} & 0.198 & 0.164 & 319.703 & \bf{0.30} & \bf{0.900}\\

\Xhline{3\arrayrulewidth}
\end{tabular}}
\vspace{-0.2cm}
\caption{Comparison of dialogue generation quality. This table presents a comparison of our model against state-of-the-art multimodal large language models, all fine-tuned on the testing dataset specifically for multimodal dialogue generation tasks. The evaluation focuses on semantic accuracy and emotional congruence of the generated response. * denotes AV-EmoDialog trained without emotion information  }
\label{table:1}
\vspace{-0.2cm}
\end{table*}

\section{Results}
\subsection{Automatic Evaluation of Dialogue Generation}
We quantitatively compare the dialogue generation results in Table 1. Our experiments were conducted on the test rare split of the MultiDialog dataset. For each conversation, we randomly selected four turns from the assistant's side. Compared with state-of-the-art multimodal large language models, our method significantly improves the EmoBERT score, which assesses the emotion alignment of the generated responses with the ground truth emotion. In addition, the semantic scores were also boosted, as evidenced by the highest scores in BLEU-1, BLEU-4, and METEOR. The high semantic performance confirms that our method generates a contextually coherent response. This result is particularly notable given that other comparison methods rely on text input, a format with which large language models are trained extensively. Compared with the cascaded baseline, which stacks a high-performing ASR, Emotion Recognition (ER) model, and LLM, our method does not require separate conversion of input speech into text and emotion labels. Nevertheless, our model achieves superior emotion alignment and comparable semantic performance, highlighting the efficiency of our end-to-end approach. Finally, AV-EmoDialog also demonstrates the highest diversity scores. These quantitative results suggest that our approach to systematically exploiting communicative cues from audio-visual inputs not only enhances emotional alignment but also improves the overall semantic quality and diversity of the generated responses. This result implies that having a better emotional context enables the model to produce responses that are more attuned to the user's state and dialogue context. 

\begin{table}
	\renewcommand{\arraystretch}{1.2}
	\renewcommand{\tabcolsep}{2.5mm}
\centering
\resizebox{0.95\linewidth}{!}{
\begin{tabular}{ccccc}
\Xhline{3\arrayrulewidth}
\multirow{2}{*}{\textbf{Method}}  &  \multirow{2}{*}{\textbf{Fleuncy}} &  \multirow{2}{*}{\makecell{\textbf{Emotional}\\ \textbf{Context}}}  &  \multirow{2}{*}{\textbf{Empathy}} \\ \\
\hline 
% {LLaVA-NeXT-img \cite{liu2024llava}} & 1.61 & 1.27 & 1.2 \\

{LLaVA-NeXT-vid \cite{liu2024llava}} &  4.32 & 1.46 & 1.26 \\

{SpeechGPT \cite{zhang2023speechgpt}}  &  3.83 & 1.56 & 1.41  \\

{Cascade} &   4.40 & 2.29 & 1.99 \\

\hdashline 
\textbf{AV-EmoDialog} &  \bf{4.64} & \bf{2.45} & \bf{2.06}  \\

\Xhline{3\arrayrulewidth}
\end{tabular}}
\vspace{-0.2cm}
\caption{\label{table:5} GPT-4 evaluation of the response in terms of fluency, emotional context, and empathy. The GPT-4 was instructed to evaluate against each criterion from 0 to 5 considering the dialogue history.}
\label{table:4}
\vspace{-0.3cm}
\end{table}

\begin{table}
	\renewcommand{\arraystretch}{1.2}
	\renewcommand{\tabcolsep}{2.7mm}
\centering
\resizebox{0.68\linewidth}{!}{
% \begin{tabular}{ccccc}
% \Xhline{3\arrayrulewidth}
% \multirow{2}{*}{\textbf{Method}}  &  \multirow{2}{*}{\textbf{Fleuncy}} &  \multirow{2}{*}{\makecell{\textbf{Emotional}\\ \textbf{Context}}}  &  \multirow{2}{*}{\textbf{Empathy}} \\ \\
% \hline 
% % {LLaVA-NeXT-img \cite{liu2024llava}} & 1.61 & 1.27 & 1.2 \\

% {LLaVA-NeXT-vid \cite{liu2024llava}} &  4.32 & 1.46 & 1.26 \\

% {SpeechGPT \cite{zhang2023speechgpt}}  &  3.83 & 1.56 & 1.41  \\

% {Cascaded} &   4.40 & 2.29 & 1.99 \\

% \hdashline 
% \textbf{AV-EmoDialog} &  \bf{4.64} & \bf{2.45} & \bf{2.06}  \\

% \Xhline{3\arrayrulewidth}
% \end{tabular}
\begin{tabular}{cc}
\hline
\textbf{Method} & \textbf{Preference (\%)} \\ \hline

{LLaVA-NeXT-vid \cite{liu2024llava}} & 34.39 \\

{SpeechGPT \cite{zhang2023speechgpt}} & 10.19 \\

Cascade & 13.38 \\

\hdashline 
\textbf{AV-EmoDialog} & 42.04 \\ \hline
\end{tabular}}
\vspace{-0.2cm}
\caption{\label{table:3} Human evaluation of the response in terms of relevance, coherence, contextual appropriateness, emotional coherence, and engagement. Participants were instructed to read the responses of each model's generation and rank them 1st to 4th.}
\label{table:5}
\vspace{-0.2cm}
\end{table}

\subsection{GPT Evaluation}
We incorporated GPT-4\footnote{\url{https://chatgpt.com/?model=gpt-4}} evaluation to further assess the generated response in Table 2.
Since EmoBERT only assesses alignment with ground truth emotional labels, a more thorough evaluation was necessary to understand the model's ability to address emotions. Henceforth, we evaluated two aspects of emotional intelligence using GPT-4: empathy which is how well the response identifies and reflects the emotion of the user, and emotional context which is whether the emotional tone of the response is suitable for the given context. The high ratings in emotional intelligence particularly suggest that our system excels at interpreting and responding to the emotional states conveyed by users.

Also, although our system directly interprets audio-visual as the input, it has even superior fluency compared to other methods that rely on textual inputs. This suggests that using audio-visual modality to directly process both linguistic and non-linguistic content not only enhances emotional understanding but also linguistic capability. 

%#################################################
\begin{figure}[htbp]
	\begin{minipage}[b]{\linewidth}
		\centering		
            \centerline{\includegraphics[width=8cm]{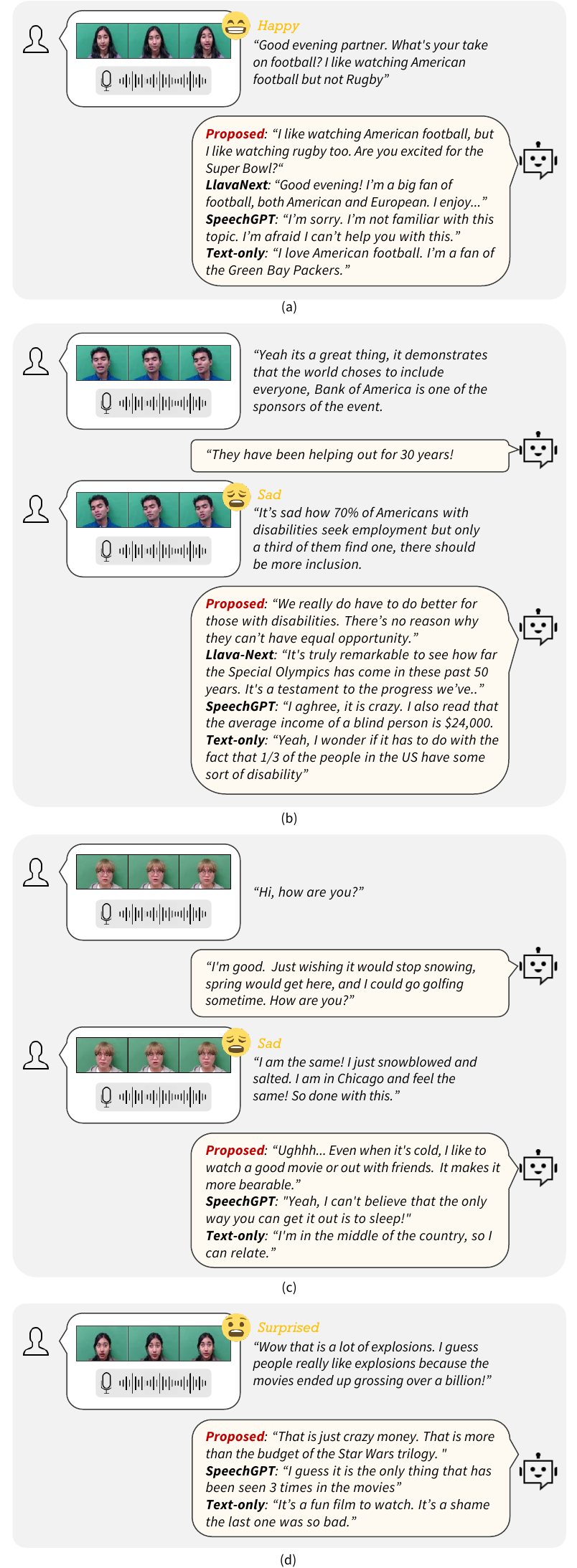}}
	\end{minipage}
        \vspace{-0.5cm}
	\caption{Generation results from our AV-EmoDialog, which processes audio-visual input from users and outputs textual responses. For clarity, the transcriptions of the audio-visual inputs are provided alongside. Below, we compare with those generated by comparison methods.}
	\label{fig:3}
  \vspace{-0.8cm}
\end{figure}
%##################################################

\subsection{Human Evaluation}
We conducted a human evaluation through Prolific\footnote{\url{https://www.prolific.com/}} to compare the quality of responses generated by various dialogue models in Table 3. 10 gathered English-speaking participants were instructed to rank generated responses from four different models according to predefined criteria: relevance, coherence, contextual appropriateness, emotional coherence, and engagement. Table 3 shows the percentage with which each model was ranked first. The results indicate that AV-EmoDialog was most frequently selected as the top-performing model, which coincides with our automatic and GPT evaluations. 

\subsection{Analysis on Audio-Visual Emotion-Aware Response}
We analyzed the emotional context of response enhanced by AV-EmoDialog in Figure 4. In (a), the user speaks in a cheerful tone with a happy smile, asking the AI about its preference for football. The proposed method effectively recognizes the user's happy emotion through audio-visual cues, which would not have been possible with text input alone. AV-EmoDialog responds accurately to the happy emotional tone of the user by saying "Are you excited for the Super Bowl?". On the other hand, the baseline methods fail to reflect the user's emotional tone, resulting in less resonant responses such as "I'm sorry. I'm not familiar with this topic" or "Good evening! I'm a big fan of football.". In (c), the user talks with a sad and depressed face about how he feels stuck in the snow. Although it may sound like the user is frustrated from the text alone, the proposed method most accurately empathizes with the user's sadness, responding with "Ughh...", and suggests one way to get out of the bad feeling.  In contrast, baseline methods misinterpret the emotional context, responding with surprise or changing the topic. Likewise in (d), the user expresses surprise about the explosion through face, tone, and the content itself. While the baseline methods moves on to a different aspect of the topic, the proposed method mirrors the user's surprise and engages further in the same emotional tone, by responding "That is just crazy money... That is more than the budget of the Star Wars trilogy.". In each example, the proposed method effectively recognizes the user's emotion from the speech tone and facial expressions, providing the most emotionally appropriate responses that would have been difficult with pure text-only input. Please refer to appendix for more generation results. Note that we provide responses in speech using an off-the-shelf TTS model\footnote{\url{https://github.com/2noise/ChatTTS}} for demonstration.

\begin{table*}
	\renewcommand{\arraystretch}{1.2}
	\renewcommand{\tabcolsep}{3.5mm}
\centering
\resizebox{0.94\linewidth}{!}{
\begin{tabular}{cccccccccc}
\Xhline{3\arrayrulewidth}
        \multirow{2}{*}{\textbf{Module}} & \multirow{2}{*}{\makecell{\textbf{Training}\\\textbf{Objectives}}} & 
        \multicolumn{1}{c}{\textbf{EMR}} & 
        \multicolumn{4}{c}{\textbf{Semantic}} & \textbf{Emotion} \\ 
        \cmidrule(lr){3-3} \cmidrule(lr){4-7} \cmidrule(lr){8-8}
        & & \textbf{Mean} & 
        \textbf{BLEU-1} & \textbf{BLEU-4} & \textbf{ROUGE} & \textbf{METEOR} & \textbf{EmoBERT} \\
        \hline 
\multirow{2}{*}{\makecell{Speech Encoder}} & ASR & -- &  0.185 & 0.0185 & 0.185 & 0.161 & 0.13 \\
& {ASR + EMR} & 0.56 &  \bf{0.212} & \bf{0.0293} & \bf{0.216} & \bf{0.179} & \bf{0.21}\\ 
\hline
\multirow{2}{*}{\makecell{Face Encoder}} & {EMR} & 0.15 & \bf0.219 & \bf{0.0355} & \bf0.222 & \bf0.166  & {0.24}   \\
& {EMR + EMD} & \bf0.26 &  {0.193} & 0.0307 & {0.198} & {0.164} & \bf0.30 \\
\Xhline{3\arrayrulewidth}\end{tabular}}
\vspace{-0.2cm}
\caption{\label{table:5} Ablation study of speech encoder and face encoder based on training objectives. `trans' indicates transcription, `emo' indicates emotion labels, and `des' denotes description of the facial expression. This table illustrates how incorporating emotion with distinct training objectives affects the performance of each encoder and enhances overall dialogue generation quality. Results with superior performance within the ablations for each respective encoder are highlighted in bold. }
\label{table:2}
\vspace{-0.2cm}
\end{table*}
\begin{table}
	\renewcommand{\arraystretch}{1.2}
	\renewcommand{\tabcolsep}{1.3mm}
\centering
\resizebox{0.99\linewidth}{!}{
\begin{tabular}{ccccccc}
\Xhline{3\arrayrulewidth}
\multirow{2}{*}{\makecell{\textbf{Input}\\ \textbf{Modality}}}
& \multicolumn{4}{c}{\centering \textbf{Semantic}} & \multicolumn{2}{c}{\centering \textbf{Emotion}} \\ 
\cmidrule(lr){2-5} \cmidrule(lr){6-7} 
 & \textbf{BLEU-4} & \textbf{ROUGE} & \textbf{METEOR} & \textbf{PPL} & \textbf{EmoBERT}  \\
\hline 
T &  0.0355 & 0.222 & 0.166 & 255.60 & 0.15 \\
A & 0.0184 & 0.222 & 0.156 & 369.56 & 0.23\\ 
% A/V & 0.212& 0.0293 & 0.216 & \bf{0.179} & 0.21 \\ 
A+V & 0.0307 & 0.198 & 0.164 & 319.70 & 0.27 \\ 
\Xhline{3\arrayrulewidth}
\end{tabular}}
\vspace{-0.2cm}
\caption{\label{table:5} Dialogue generation performance with different modality input, including text, audio, and audio-visual.}
\label{table:3}
\vspace{-0.2cm}
\end{table}

\subsection{Effectiveness of Exploiting Emotion in Audio-Visual Dialogue}
We tested AV-EmoDialog with emotion excluded from the last stage of training on the audio-visual dialogue task. The results are shown in the second to last row in Table 1. There was a huge drop in emotion performance - from 0.3 to 0.2, which suggests that having the emotion label in the dialogue allows the model to keep track of the understanding of the emotional context. 

We evaluated the training schemes used for the speech and face encoders in Table 3. We observed that incorporating the detailed descriptions of the facial expressions greatly increased the emotion recognition performance of the face encoder by 70\%. Also, the emotion recognition performance of the speech encoder is higher than from the face. This indicates that vocal expressions, often being more expressive and straightforward, can capture emotional nuances more effectively than visual expressions. But since they distinctively have their effects, we leverage both cues to boost the performance in the dialogue generation task.  

In the dialogue generation task, speech encoder trained with emotion recognition objective has better performance (rows 1 and 2). This demonstrates that having the emotion label gives better context to generate a response that is more emotionally and semantically aligned. Furthermore, the face encoder, trained with detailed facial descriptions, also showed enhanced emotion recognition capabilities, which impacted its dialogue generation performance (rows 3 and 4). While there was a notable increase in the EmoBERT score, a slight decrease in semantic scores was observed. This reduction may be due to the complexities introduced by the detailed emotion descriptions, which could complicate the dialogue modeling. Nevertheless, its effect on emotion performance is evident, and semantic performance remains competitive, surpassing that of the baselines.

\subsection{Comparison of Audio, Visual, and Text as Input for Emotion-aware Dialogue}
In Table 5, we conducted experiments with various modality inputs to validate the efficacy of each modality in dialogue generation. The text-only input has the highest semantic quality overall, which can largely be attributed to the fact that the underlying llm has been extensively trained on textual data. Yet, utilizing both audio-visual modalities was able to achieve on-par semantic performance while achieving superior emotional intelligence. 

Another finding is that using both audio-visual modalities results in enhanced semantic and emotional performance compared to using audio modality alone. This improvement underscores the value of visual cues such as facial expressions and eye contact, which provide complementary information to audio cues such as pitch and tone of voice. Our approach reflects natural human communication, where verbal cues are typically supported and enriched by non-verbal expressions, creating a more holistic and effective interaction.
% that our approach of exploiting emotional cues within each modality effectively enhances the model’s emotional responsiveness, providing a more comprehensive understanding of the context. 
% Our findings suggest that our AV-EmoDialog method of integrating audio-visual cues effectively enhances the model’s emotional responsiveness.

\section{Conclusion and Limitation}
The machine's ability to recognize the user's emotional state has great implications for human-machine communications; a virtual assistant that recognizes frustration can offer more targeted help, a customer service bot equipped with empathy can better handle complaints, and therapeutic service can comprehensively diagnose patients. Our proposed AV-EmoDialog contributes to the field of emotion-aware dialogue systems by exploiting linguistic and non-linguistic information directly from audio-visual inputs. AV-EmoDialog relies on audio-visual input without additional text input to optimize the system end-to-end for emotion-aware audio-visual dialogue. By leveraging fine-grained facial descriptions, AV-EmoDialog not only aligns more closely with human communication patterns but also sets a new standard for the design of empathetic and effective dialogue systems. Our extensive experiments confirm that AV-EmoDialog surpasses existing baseline models in both emotional and semantic dimensions.

To further explore the potential of our proposed model, which is designed to handle emotional interplays in dialogues, it would be beneficial to have a more diverse audio-visual dataset from real-world scenarios, where emotional interactions are more prevalent. Also, the future research direction would be to generate the speech in end-to-end. By generating speech that reflects the emotional tone of the response, it would be possible to create more immersive conversations for users. 
For ethical considerations, as the system directly utilizes audio-visual user data, it is crucial to ensure the privacy and security of sensitive information. 

% The limitation of the work is that it has been tested exclusively on Multidialog because it is the only available large-scale dialogue dataset covering audio, visual, and text. During our experimentation, we observed that the dataset predominantly focuses on knowledge-centric interactions, which do not extensively require emotional exchanges between speakers. 

{
    \small
    \bibliographystyle{ieeenat_fullname}
    \bibliography{main}
}

\clearpage
\setcounter{page}{1}
\maketitlesupplementary

\section{Evaluation Metrics}
\noindent \textbf{BLEU} \cite{post2018call} evaluates the fluency and adequacy of generated responses based on n-gram overlap. A higher BLEU score indicates a more natural and engaging dialogue model. We measure BLEU-1 and BLEU-4 score where BLEU-1 measures the overlap at unigram and BLEU-4 considers up to 4-grams, focusing more on the contextual and syntactical relationships between words. 

\noindent \textbf{PPL} \cite{bengio2000neural} measures how well a language model predicts the generated response. A lower perplexity indicates that the model is more confident and accurate in predicting the next word, suggesting higher quality in generating coherent and contextually relevant responses.

\noindent \textbf{METEOR} \cite{banerjee2005meteor} assesses the quality of generated response by computing the alignment-based precision and recall between the generated output and the ground truth, considering synonyms and paraphrases. 

\noindent \textbf{ROUGE} \cite{lin2004rouge} measures the overlap of n-grams between the generated output and a set of reference outputs based on recall (\ie how much of the reference is captured by the output).

\noindent \textbf{EmoBERTscore} \cite{zhang2019bertscore} assess the semantic similarity between the generated response emotion and the reference response emotion. We utilize the BERT model specifically fine-tuned on GoEmotions \cite{demszky2020goemotions} dataset by Google, which includes 58k Reddit comments labeled for 27 emotion categories including neutral. This model is adept at detecting a broad range of emotions. The features extracted from the BERT model are used to calculate the similarity distance between the ground truth and the generated response.  

\section{GPT evaluation} We ran GPT evaluation on the following criteria; coherence, fluency, and emotional intelligence. Fluency evaluates the grammatical correctness, smoothness, and natural flow of the response. The response should read as if it were spoken or written by a fluent language user. Emotional context evaluates how appropriate the emotion conveyed in response is within the context of the dialogue history. Empathy evaluates how well the response demonstrates an awareness of the user's feelings. We instruct GPT to provide a score in the range of 0 to 5 with regard to the three criteria. The prompt given to the GPT is shown in Figure~\ref{fig:6}.

\section{Human evaluation}  
We conducted a human evaluation through Prolific\footnote{\url{https://www.prolific.com/}} to compare the quality of responses generated by various dialogue models. Ten English-speaking participants were recruited and tasked with ranking the responses generated by four different models based on predefined criteria: relevance, coherence, contextual appropriateness, emotional coherence, and engagement. Detailed instructions provided to the participants are shown in  Figure~\ref{fig:human_evaluation_details}.

\section{Datasets}
\noindent \textbf{Common Voice} \cite{ardila2019common} is a large-scale multilingual speech-text corpus collected from volunteers around the world. It contains 2,615 hours of English speech from 92,325 voices with diverse genders, ages, and accents. 

\noindent \textbf{GigaSpeech} \cite{chen2021gigaspeech} an English speech-text corpus from various sources such as podcasts, audiobooks, and YouTube, accompanied by accurate transcriptions. We use a large subset which is 2,500 hours.  It is the common benchmark used for speech processing tasks.

\noindent \textbf{LibriSpeech} \cite{panayotov2015librispeech} is approximately 1,000 hours of English speech derived from audiobooks in the LibriVox project. 

\noindent \textbf{MultiDialog} \cite{park2024let} is a large-scale audio-visual dialogue dataset, consisting of approximately 370 hours of 9,000 dialogues between 6 pairs of speakers. It is based on the Topical Chat dataset which is a knowledge grounded human-human conversation covering 9 broad topics. There are emotion annotations for each utterance. It is the only available large-scale audio-visual dialogue dataset to date.

\noindent \textbf{RAVDESS} \cite{livingstone2018ryerson} is an emotional audio-visual data consisting of 1,4440 files from 24 actors (12 female and 12 male) speaking in a natural North American accent and in six different emotions (happy, sad, angry, fearful, surprise, disgust, and calm) with two different intensity levels (normal and strong)

\noindent \textbf{CREMA-D} \cite{cao2014crema} is an emotional audio-visual data consisting of 7,442 audio-visual clips from 91 actors (48 male and 43 female) between the ages of 20 to 74 from various races (African American, Asian, Caucasian, Hispanic, and unspecified). They speak in six different emotions (anger, disgust, fear, happy, neutral, and sad) in four different emotion levels (low, medium, high, and unspecified).

%#################################################
\begin{figure}[ht]
	\begin{minipage}[b]{\linewidth}
		\centering		
            \centerline{\includegraphics[width=9cm]{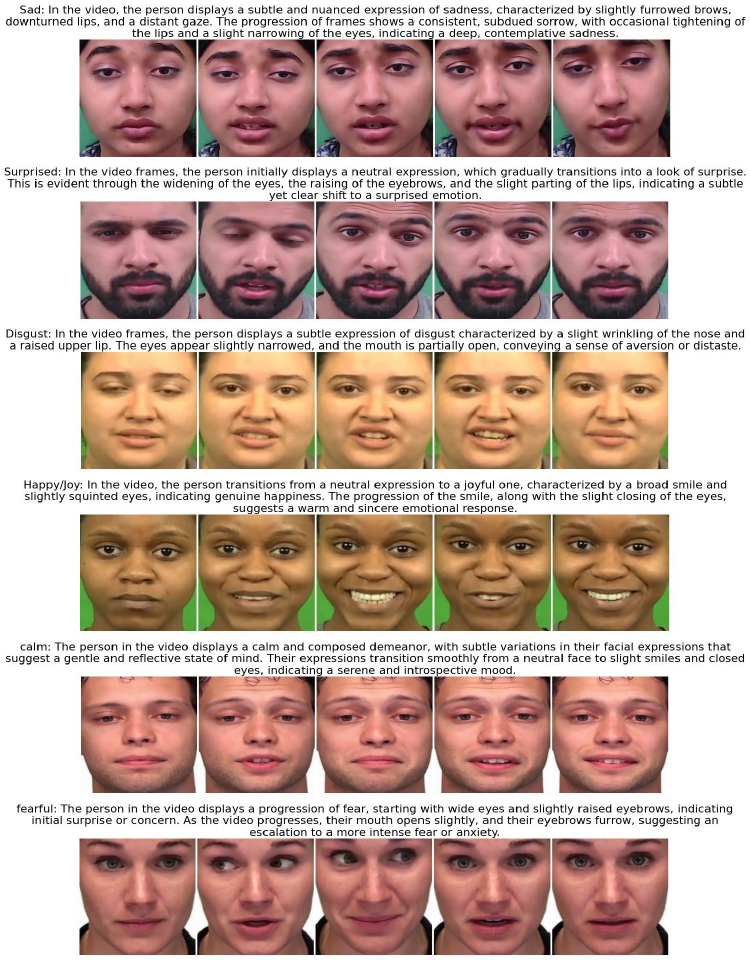}}
	\end{minipage}
        \vspace{-0.5cm}
    \caption{Facial descriptions generated for training using GPT.}
	\label{fig:5}
  \vspace{-0.3cm}
\end{figure}
%##################################################

\section{Extracting Facial Descriptions}
We use GPT to annotate the face video with rich facial descriptions relevant to emotions. The prompt given to the GPT is: 
\textit{These are the frames in a video. In the video, a person expresses emotion through facial expressions. Describe the subtle and fine-grained emotion precisely as you see in the video in two sentences.} We only annotate a subset of RAVDESS, CREMA-D, and Multidialog for training. We will open-source the annotations. The samples are shown in Figure~\ref{fig:5}. The generated descriptions provide detailed information of facial dynamics associated with emotions, capturing the progression and intensity of the emotion—nuances that cannot be conveyed by a single-word emotion category.

\section{More Generated Results}

We have included more generation results in Figure 8, which presents a comparison of the generation results from our proposed method against those from LLaVA-Next~\cite{liu2024llava}, SpeechGPT~\cite{zhang2023speechgpt}, and the Cascade model. Our AV-EmoDialog directly audio-visual as the input and generated more emotion-aware response, guided by the audio-visual input. 

In Figure 8(a), AV-EmoDialog responds with a semantically coherent answer, maintaining the context of FDA burning tons of books. In contrast, LLaVA-Next's response deviates to an unrelated topic involving Ted Koppel's controversy. While SpeechGPT and Cascade stay on topic, their responses lack emotional awareness. From the text user input of "Yeah, I think it's unfair how the FD burns 6 tons of books", it's unclear what the user's emotion is. But since AV-EmoDialog can better recognize emotion through audio-visual input, it considers the emotion and replies, "That is awful.".

In Figure 8(b), AV-EmoDialog demonstrates its ability to empathize with the user's sadness by responding, "It's sad to see things like this." This response reflects a direct acknowledgment of the user's emotional state. LLaVA-Next also appears to empathize with the user but takes a more pragmatic approach, suggesting detailed ways to address the issue, which could be explored further in a broader context. SpeechGPT fails to align with the emotional context, responding, "I think they should be happy though." The Cascade model remains contextually relevant but pays less attention to incorporating the user's emotional tone into its response. 

In Figure 8(c), AV-EmoDialog effectively maintains the context about Nintendo and builds on the previous turn by introducing another surprising fact. LLaVA-Next initially empathizes with the user's emotion but diverges from the topic, discussing several unrelated aspects of Nintendo. The Cascade model fails to consider the user's prior input saying, "I wonder if you can use them on other consoles, like the Nintendo switch.". SpeechGPT stays on topic, continuing the discussion about the fact, but it does not capture or reflect the user's surprised emotion.

In Figure 8(d), the proposed method responds appropriately regarding how only a third of Americans with disabilities find employment by saying "We really do have to do better for those with disabilities~". Besides the cascade method, both LLaVA-Next and SpeehGPT diverge from the topic, failing to address the issue of high unemployment rates among disabilities or to continue the conversation meaningfully.

In Figure 8(e), the discussion focuses on Disney, with the user expressing surprise about Disney's Big Hero 6 being rendered using a powerful supercomputer. SpeecghGPT completely misses the topic, LLaVA-Next talks about Disney characters, and Cascade reflects the surprised emotion by fails to continue on with the conversation. In contrast, our proposed model not only mirrors the user's surprised emotion with an opening like "Wow" but also advances the discussion by asking a related question about the capabilities of such supercomputers.

From these samples, it is evident that AV-EmoDialog is more emotion-aware and generates responses that are more coherent compared to the other models. 
This underlines the efficacy of AV-EmoDialog in handling complex dialog scenarios where emotional context influences the response quality. We have attached a demo video of generated samples from AV-EmoDialog. For better demonstration, we have produced speech from the generated response using a TTS module.

%##################################################

%#################################################
\begin{figure*}[htbp]
	\begin{minipage}[b]{\linewidth}
		\centering		
            \centerline{\includegraphics[width=13cm]{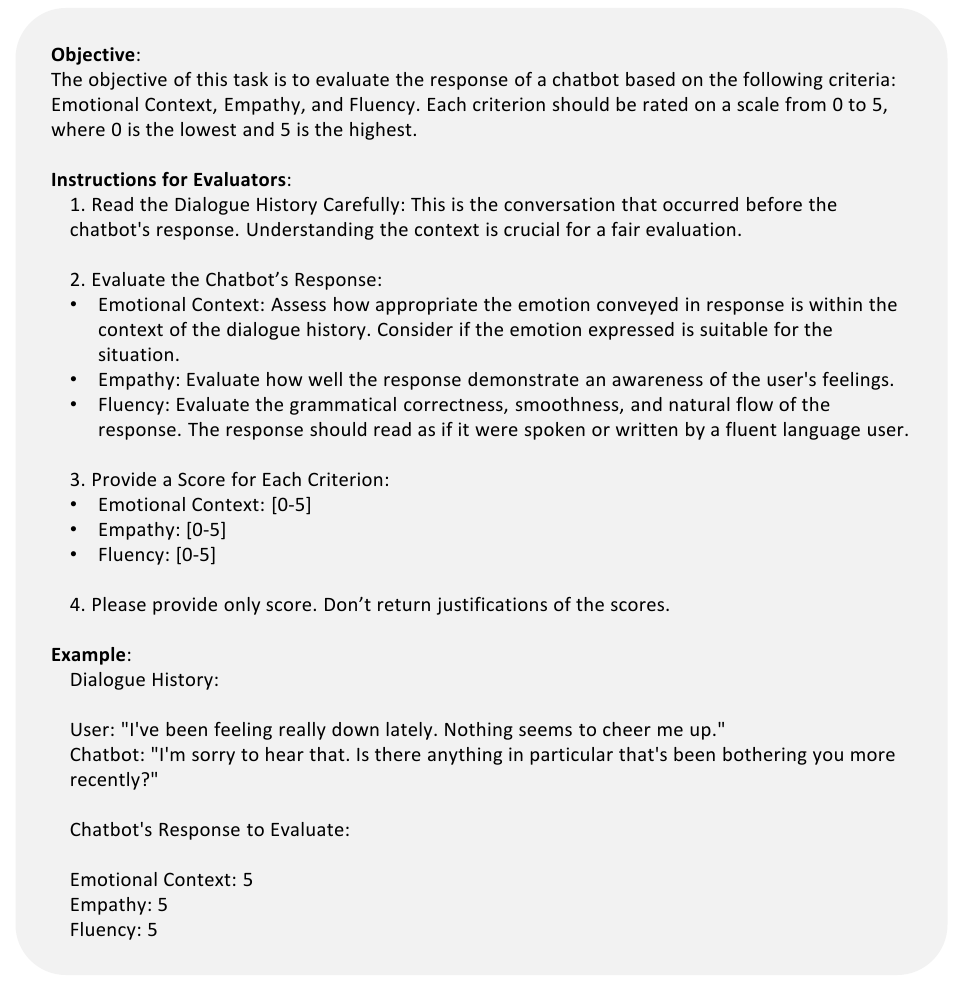}}
	\end{minipage}
        \vspace{-0.5cm}
    \caption{GPT Evaluation prompt to evaluate the response in terms of emotional context, empathy, and fluency, .}
	\label{fig:6}
  \vspace{-0.3cm}
\end{figure*}
%##################################################

%#################################################
\begin{figure*}[htbp]
	\begin{minipage}[b]{\linewidth}
		\centering		
            \centerline{\includegraphics[width=12cm]{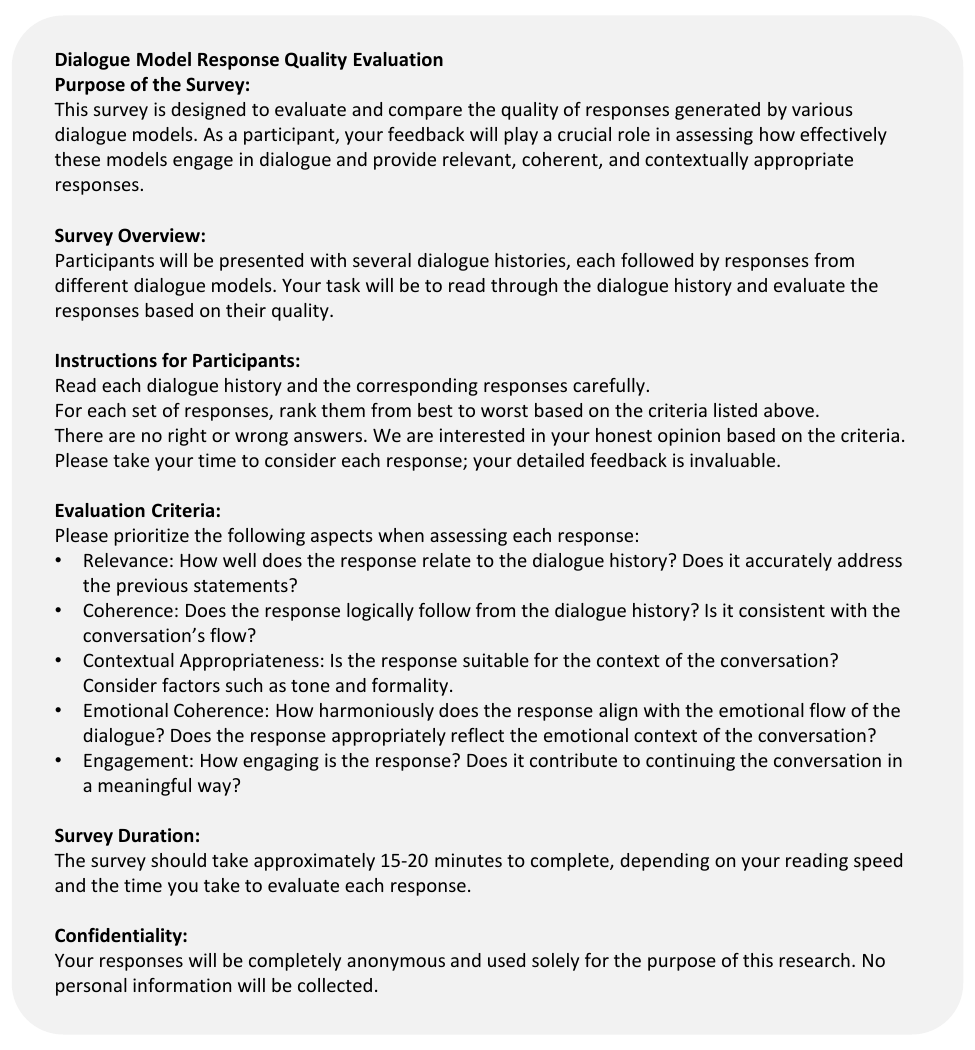}}
	\end{minipage}
        \vspace{-0.5cm}
    \caption{Human evaluation guideline for participants.}
	\label{fig:human_evaluation_details}
  \vspace{-0.3cm}
\end{figure*}
%##################################################

%#################################################
\begin{figure}[htbp]
	\begin{minipage}[b]{\linewidth}
            \begin{minipage}[t]{\linewidth}
		  \centering		   
            \centerline{\includegraphics[width=8cm]{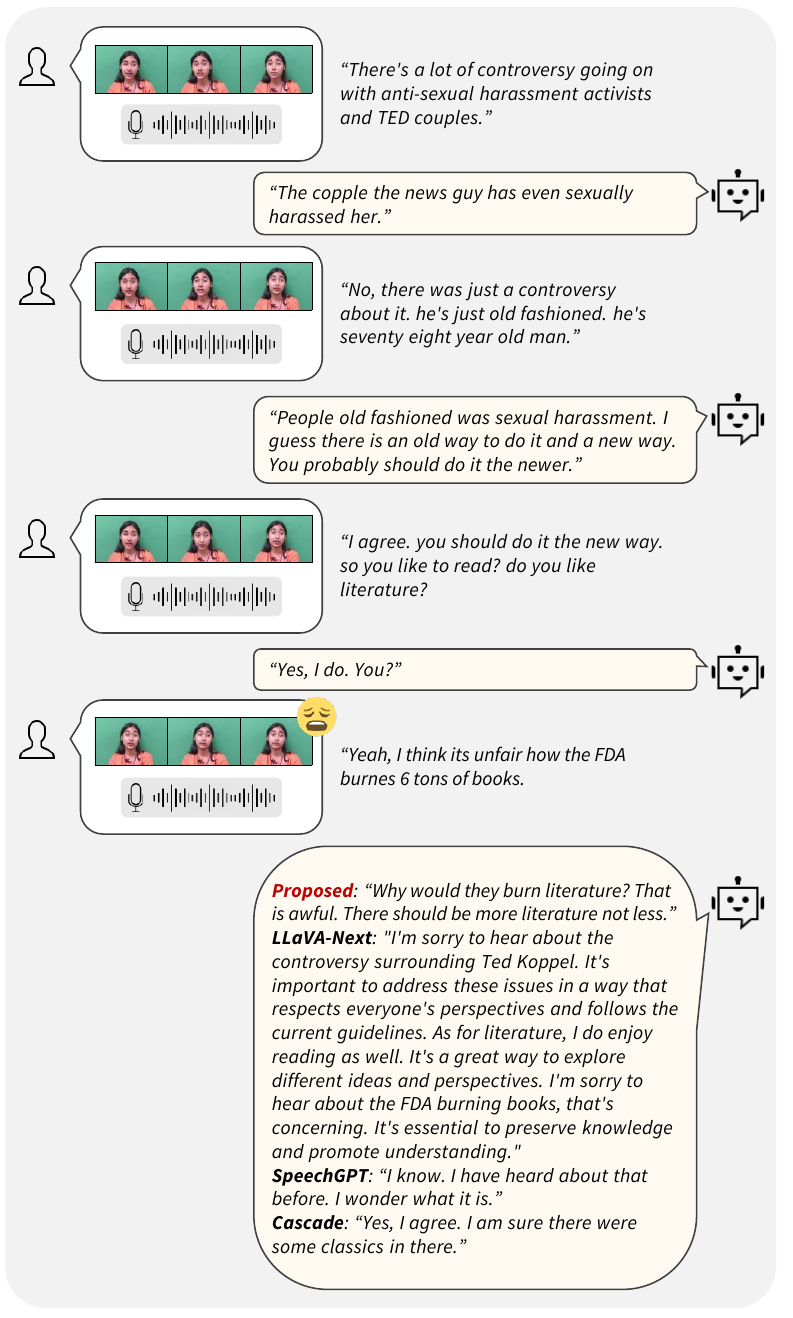}}
            \caption*{Figure 8 (a).}
            \end{minipage}\\
            \begin{minipage}[t]{\linewidth}
            \centering	
            \vspace{0.3cm}
            \centerline{\includegraphics[width=8cm]{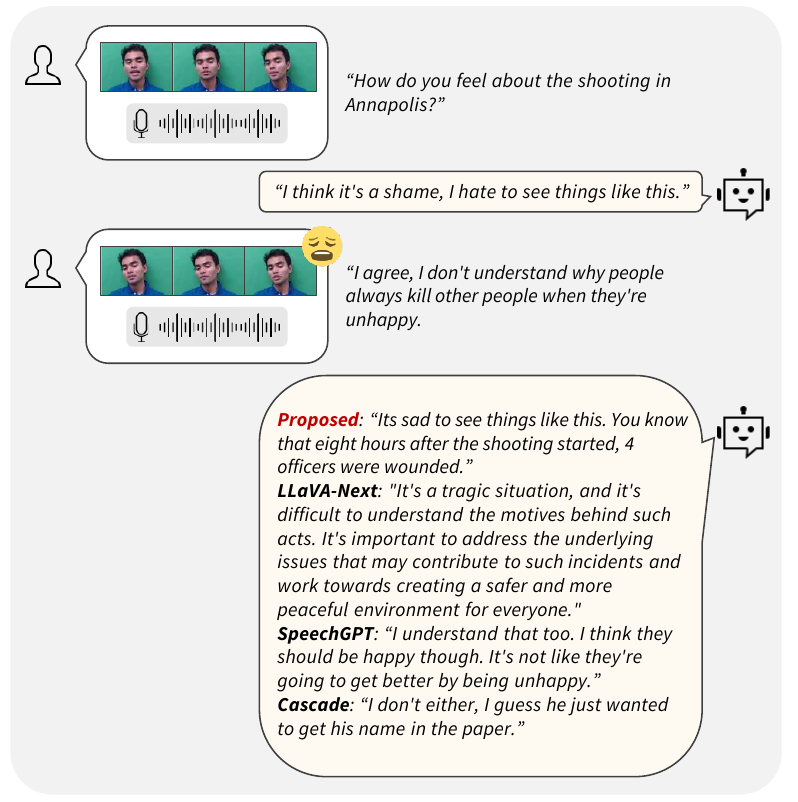}}
            \caption*{Figure 8 (b).}
            \end{minipage}
	\end{minipage}
        % \vspace{-0.5cm}
  % \vspace{-0.8cm}
\end{figure}
\begin{figure}[htbp]
	\begin{minipage}[b]{\linewidth}
            \centering		
            \centerline{\includegraphics[width=8cm]{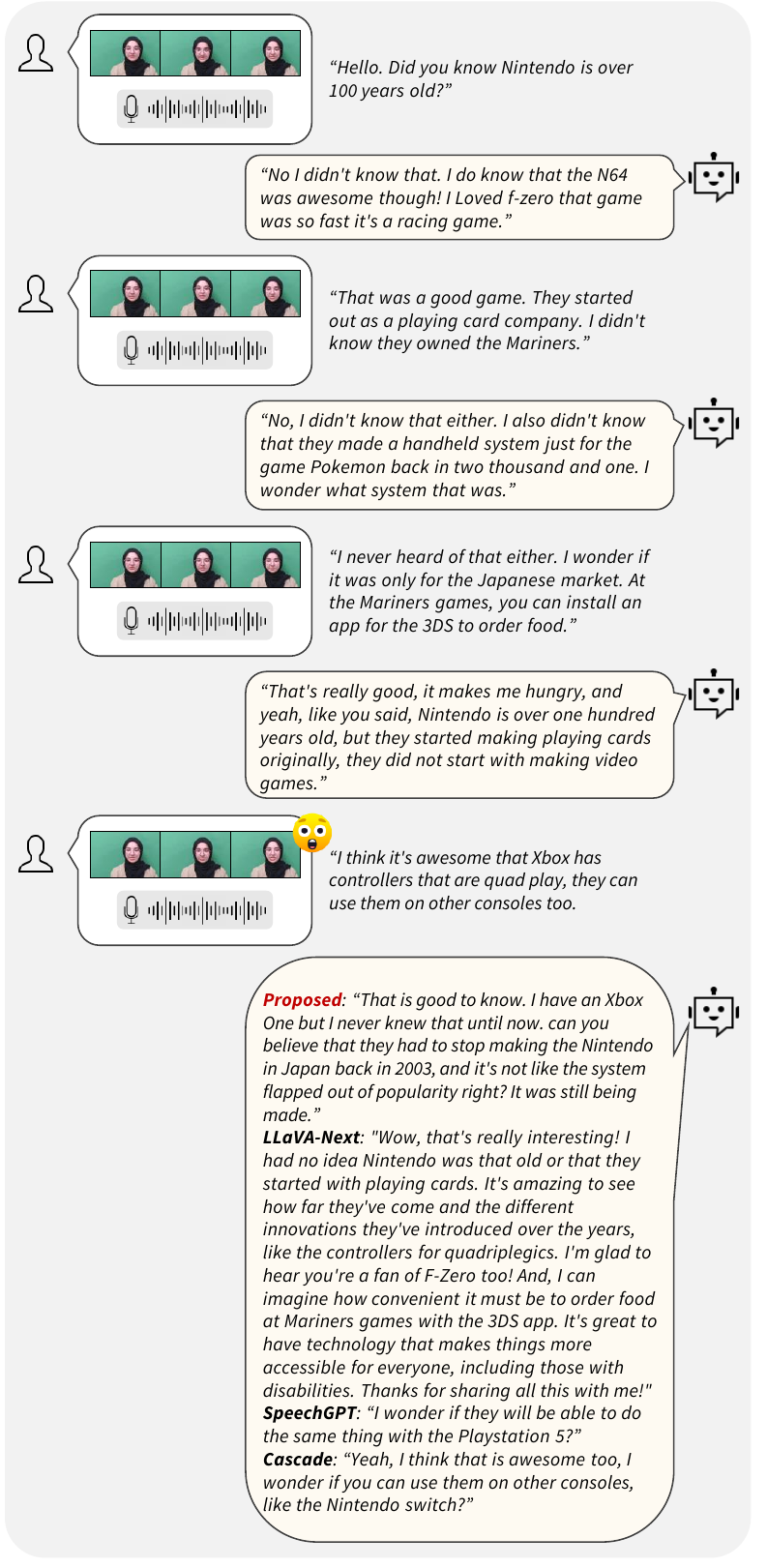}}
            \caption*{Figure 8 (c).}
	\end{minipage}
        % \vspace{-0.5cm}
  % \vspace{-0.8cm}
\end{figure}

\begin{figure*}[t]
	\begin{minipage}[t]{0.48\linewidth}
		\centering		
            \centerline{\includegraphics[width=8cm]{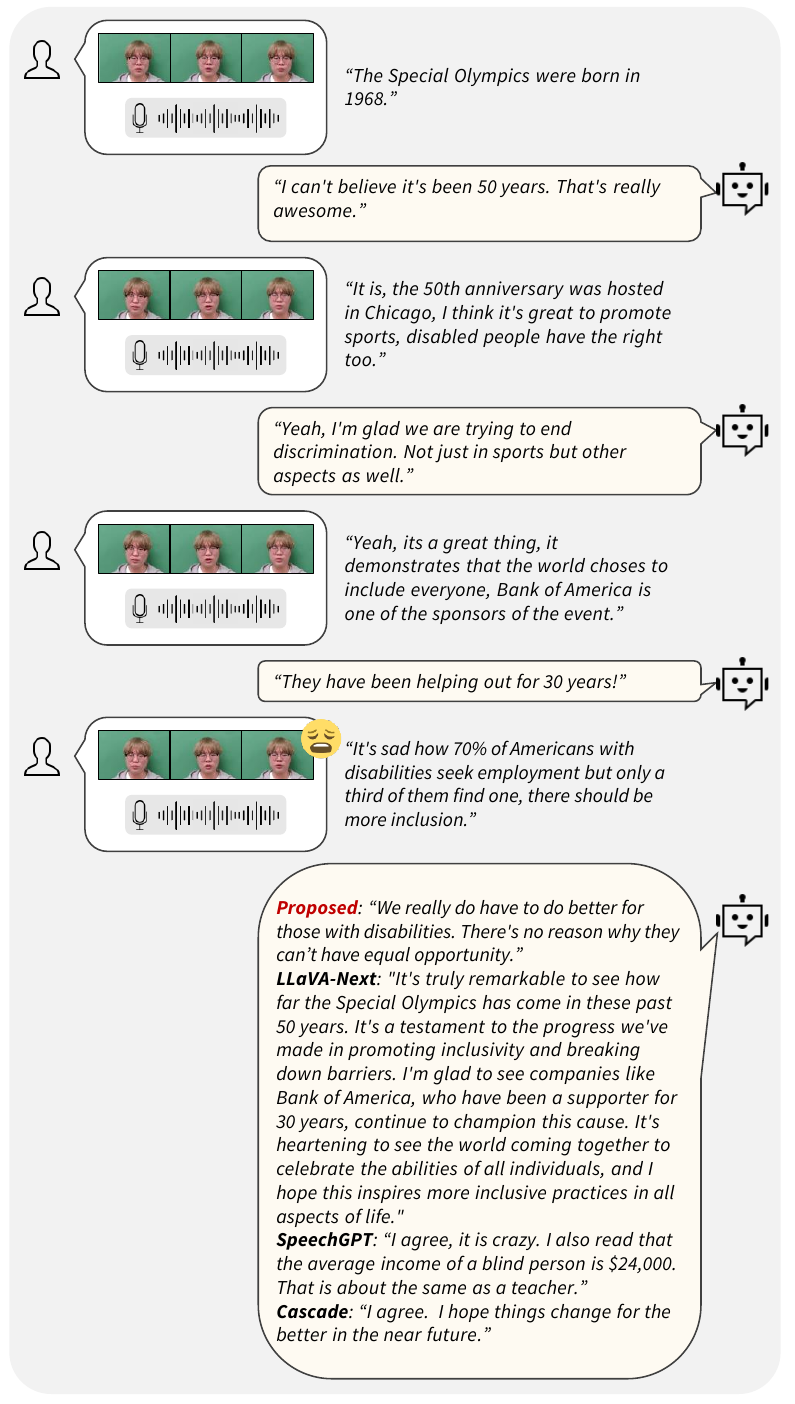}}
            \caption*{Figure 8 (d).}
	\end{minipage}\hfill
         \begin{minipage}[t]{0.48\linewidth}
		\centering		
            \centerline{\includegraphics[width=8cm]{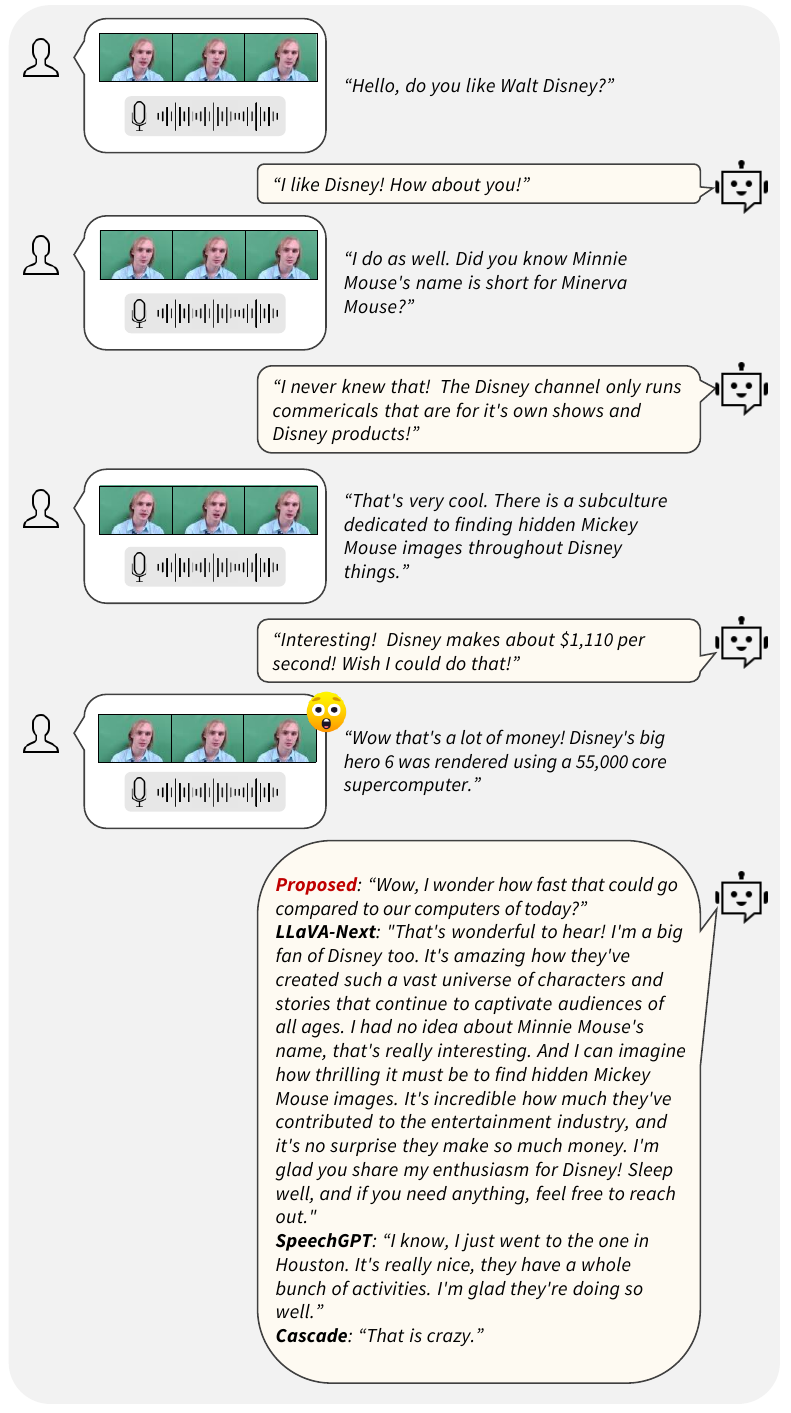}}
            \caption*{Figure 8 (e).}
	\end{minipage}
        % \vspace{-0.5cm}
  \vspace{-0.3cm}
\end{figure*}
%##################################################

\

% \section{Hard Prompts}
% \subsection{Speech Understanding Prompt}
% The speech understanding prompt is tailored to train the audio encoder on speech recognition tasks coupled with the extraction of relevant metadata. The LLM ingests speech embeddings as input and predicts the corresponding transcriptions and contextual metadata, which may include tone, speech rate, or background noise descriptions when available. This prompt underpins the model's capability to convert spoken audio into the verbal cues that the LLM can interpret. 

% \subsection{Face Video Understanding Prompt}
% The face video understanding prompt is designed to train the face encoder on visual emotion recognition tasks. The prompt directs the LLM to process the video embeddings as input and predict nuanced emotions along with detailed descriptions. To assi

% 
% To split the supplementary pages from the main paper, you can use \href{https://support.apple.com/en-ca/guide/preview/prvw11793/mac#:~:text=Delete%20a%20page%20from%20a,or%20choose%20Edit%20%3E%20Delete).}{Preview (on macOS)}, \href{https://www.adobe.com/acrobat/how-to/delete-pages-from-pdf.html#:~:text=Choose%20%E2%80%9CTools%E2%80%9D%20%3E%20%E2%80%9COrganize,or%20pages%20from%20the%20file.}{Adobe Acrobat} (on all OSs), as well as \href{https://superuser.com/questions/517986/is-it-possible-to-delete-some-pages-of-a-pdf-document}{command line tools}.

\end{document}